\documentclass[runningheads]{llncs}
\usepackage{graphicx}

\usepackage{microtype}

\usepackage{multirow}
\usepackage{amsmath}
\usepackage{booktabs}
\usepackage{amssymb}
\usepackage{algorithm}
\usepackage{makecell}
\usepackage{CJK} 
\usepackage{graphicx}
\usepackage{subfigure}
\usepackage{multirow}
\usepackage{palatino}
\usepackage{tikz}
\usepackage{float}

\usepackage{algorithmic}
\usepackage[colorlinks,
linkcolor=blue,
anchorcolor=blue,
citecolor=blue]{hyperref}

\begin{document}
%
\title{Unique Chinese Linguistic Phenomena}
%

%

\author{Shengbin Jia}

\authorrunning{S. Jia et al.}
\institute{Tongji University, Shanghai 201804, China \\
	\email{shengbinjia@tongji.edu.cn}
}
\maketitle              

\section{Introduction}

Linguistics holds unique characteristics of generality, stability, and nationality \cite{Zheng2013}, which will affect the formulation of extraction strategies and should be incorporated into the relation extraction. Chinese open relation extraction is not well-established, because of the complexity of Chinese linguistics makes it harder to operate, and the methods for English are not compatible with that for Chinese. The diversities between Chinese and English linguistics are mainly reflected in morphology and syntax. There are three differences in morphology \cite{Zheng2013,Chen2012}.

\begin{enumerate}
	\item Each word is separated by space in English; however, there is no obvious dividing mark among Chinese words.

	\item  English words vary in their tense, but Chinese words do not. The same Chinese expression may act as a different semantic function that may generate ambiguity. For example, the sentence ``\begin{CJK*}{UTF8}{gbsn}托马斯\end{CJK*}(Tomas)  \begin{CJK*}{UTF8}{gbsn}在\end{CJK*}(at)  \begin{CJK*}{UTF8}{gbsn}肯德基\end{CJK*}(KFC)  \begin{CJK*}{UTF8}{gbsn}吃\end{CJK*}(eat)  \begin{CJK*}{UTF8}{gbsn}早餐\end{CJK*}(breakfast)." can express semantics that Tomas usually eats breakfast at KFC  or Tomas will eat breakfast at KFC. The verb \begin{CJK*}{UTF8}{gbsn}吃\end{CJK*}(eat) can be declared as a habitual action by the present tense (eats), or indicates what will happen using future tense (will eat). This should be judged regarding specific contexts.
	
	\item There is only one predicate head in an English sentence, and there are many conjunctions among clauses. While predicate heads might not be unique in a Chinese sentence, multiple subordinate clauses can be connected without any conjunctions. For example, ``\begin{CJK*}{UTF8}{gbsn}乔丹\end{CJK*}(Jordan) \begin{CJK*}{UTF8}{gbsn}是\end{CJK*}(is) \begin{CJK*}{UTF8}{gbsn}美国\end{CJK*}(American) \begin{CJK*}{UTF8}{gbsn}职业\end{CJK*}(professional) \begin{CJK*}{UTF8}{gbsn}篮球\end{CJK*}(basketball) \begin{CJK*}{UTF8}{gbsn}运动员\end{CJK*}(player), \begin{CJK*}{UTF8}{gbsn}出生\end{CJK*}(born) \begin{CJK*}{UTF8}{gbsn}在\end{CJK*}(in) \begin{CJK*}{UTF8}{gbsn}纽约\end{CJK*}(New York).". There are two predicates \begin{CJK*}{UTF8}{gbsn}是\end{CJK*}(is) and \begin{CJK*}{UTF8}{gbsn}出生\end{CJK*}(born) in two clauses respectively and no conjunctions to connect the clauses. However, the  idiomatic expression in English is ``Jordan is an American professional basketball player who was born in New York.".
	
\end{enumerate}

Thus, many methods using morphological features that usually perform well in English corpus, may achieve poor results in Chinese \cite{Chen2015,Wang2011,Chen2012}. Considering the deeper linguistic characteristics combining with relation extraction to obtain a model adapted to Chinese, is one of the core research topics of this article. 

Some Chinese syntactic use is similar to English, such as the ``subject-predicate-object" structure in the example sentence ``\begin{CJK*}{UTF8}{gbsn}海森\end{CJK*}(Hassan) \begin{CJK*}{UTF8}{gbsn}上周\end{CJK*}(last week) \begin{CJK*}{UTF8}{gbsn}离开了\end{CJK*}(left) \begin{CJK*}{UTF8}{gbsn}夏威夷\end{CJK*}(Hawaii).". However, there are quite a few syntactic phenomena that are different from English, too. These should be the focus to amend the state-of-the-art English ORE systems to extract relations from Chinese text. We define three main Chinese linguistic phenomena that are categorically different from English. The analysis lays a theoretical foundation for designing the model in Section 4.

\begin{definition}Nominal Modification-Center (NMC) Phenomenon.\end{definition}

Nominal modification-center consists of the modifiers and the head word, which is a common noun. The head word may play the role of the subject or object in a sentence. Specially, when the NMC contains a named entity as a modifier, such as ``\begin{CJK*}{UTF8}{gbsn}奥巴马\end{CJK*}(Obama) \begin{CJK*}{UTF8}{gbsn}总统\end{CJK*}(president)...", ``\begin{CJK*}{UTF8}{gbsn}华盛顿\end{CJK*}(Washington) \begin{CJK*}{UTF8}{gbsn}警方\end{CJK*}(Police)...", such a situation often interferes with the extracting: Instead of the entity directly being the subject or object, the head word usually becomes the main component of the phrase. English likes to directly use an entity as the head word or with the help of prepositions, such as ``president Obama", ``the Police of Washington". We define the head word as \textbf{Pseudo-entity} to do the corresponding conversion during extracting. For example, ``\begin{CJK*}{UTF8}{gbsn}奥巴马\end{CJK*}(Obama) \begin{CJK*}{UTF8}{gbsn}总统\end{CJK*}(president) \begin{CJK*}{UTF8}{gbsn}访问\end{CJK*}(visited) \begin{CJK*}{UTF8}{gbsn}中国\end{CJK*}(China).", from which relation \begin{CJK*}{UTF8}{gbsn}访问\end{CJK*}(visited) between the Pseudo-entity \begin{CJK*}{UTF8}{gbsn}总统\end{CJK*}(president) and entity \begin{CJK*}{UTF8}{gbsn}中国\end{CJK*}(China) is easy to get. Then, we can convert and output the relation three tuple (\begin{CJK*}{UTF8}{gbsn}奥巴马\end{CJK*}(Obama), \begin{CJK*}{UTF8}{gbsn}访问\end{CJK*}(visited), \begin{CJK*}{UTF8}{gbsn}中国\end{CJK*}(China)).

\begin{definition}Chinese Light Verb Construction  (CLVC) Phenomenon.\end{definition}
CLVC is a co-occurrence that a light verb must be with nouns, the patient of the verb appears in the form of a preposition object, and the position of the preposition is flexible and changeable. 

In linguistics, a light verb carries little semantic content and typically forms a predicate with a noun \cite{Butt2003}. In English, the noun phrase locating between a light verb and a preposition is usually regarded as a light verb construction (LVC). For example, with the phrases ``is the president of..." and ``establish diplomatic relations with...", in which ``is" and ``establish" are the light verbs. Many papers have been demonstrated that unreasonable handling of LVC could cause significant uninformative extractions \cite{Etzioni2011,Qiu2014,Fader2011}.
If only the light verb is extracted as the relationship word, the relation triples, such as (\begin{CJK*}{UTF8}{gbsn}巴拿马\end{CJK*}(Panama), \begin{CJK*}{UTF8}{gbsn}建立\end{CJK*}(established), \begin{CJK*}{UTF8}{gbsn}中国\end{CJK*}(China)) from the sentence ``\begin{CJK*}{UTF8}{gbsn}巴拿马\end{CJK*}(Panama) \begin{CJK*}{UTF8}{gbsn}在\end{CJK*}(in) 2017 \begin{CJK*}{UTF8}{gbsn}年\end{CJK*}(year) \begin{CJK*}{UTF8}{gbsn}与\end{CJK*}(with) \begin{CJK*}{UTF8}{gbsn}中国\end{CJK*}(China) \begin{CJK*}{UTF8}{gbsn}建立\end{CJK*}(established) \begin{CJK*}{UTF8}{gbsn}外交\end{CJK*}(diplomatic) \begin{CJK*}{UTF8}{gbsn}关系\end{CJK*}(relations).", are probably wrong. The English extractor REVERB  \cite{Fader2011} solved the problem by syntactic constraints: the light verbal phrases must be a contiguous word sequence of  ``verb-noun-preposition".

\begin{figure}[H]
	\centering
	\subfigure[]{
		\label{fig:two:a} 
		\includegraphics[width=0.5\textwidth]{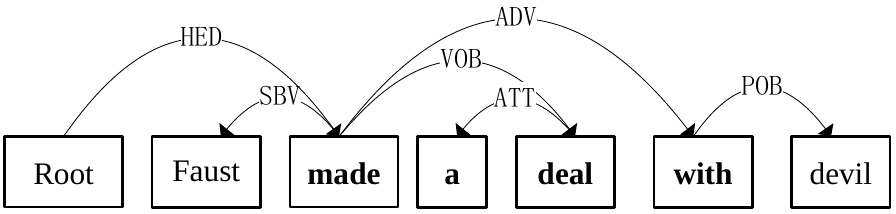}}
	\hspace{1in} 
	\subfigure[]{
		\label{fig:two:b} 
		\includegraphics[width=0.5\textwidth]{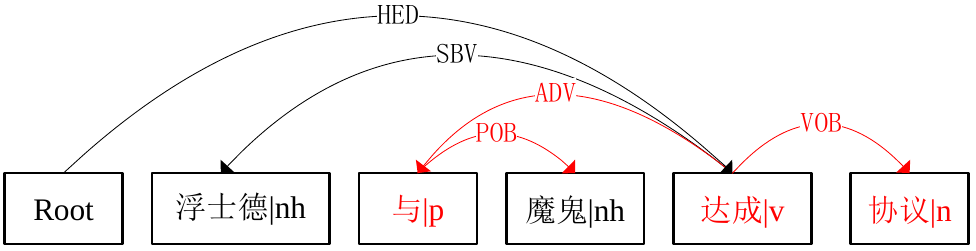}}
	\caption{The (a) graph shows the quoted sentence from Fader's paper, and it's dependency trees, bold annotation phrase is the LVC. The (b) graph presents the dependency parsing of the Chinese translation of the sentence in (a). And we mark the key construction in red.}
	\label{fig:two} 
\end{figure}

However, the word sequences of LVCs in English are quite different from Chinese. Therefore, the syntactic constraints in REVERB cannot be directly transferred to Chinese extractors. As the example in paper of \cite{Fader2011}, in ``Faust made a deal with the devil.", ``made a deal with" is an LVC and  so is the predicate phrase. The example results in the relation triple (Faust, made a deal with, the devil). Translating the sentence into Chinese gives us \begin{CJK*}{UTF8}{gbsn}浮士德\end{CJK*}(Faust) \begin{CJK*}{UTF8}{gbsn}与\end{CJK*}(with) \begin{CJK*}{UTF8}{gbsn}魔鬼\end{CJK*}(devil) \begin{CJK*}{UTF8}{gbsn}达成\end{CJK*}(made) \begin{CJK*}{UTF8}{gbsn}协议\end{CJK*}(a deal). However, as is vividly shown in Figure~\ref{fig:two}, we can find that the sequence of ``verb-noun-preposition" will no longer be maintained since the position of the preposition is more flexible in CLVC.

We propose to employ the dependency structure to solve the trouble of CLVC during relation extracting. It is found that dependency parsing can well represent the law of CLVC. Regardless of the position of the preposition in the sentence, the preposition and entity are combined into an adverbial phrase and appear in dependency parsing. The preposition depends on the verb by ADV, and there is an entity object that depends on the preposition by POB. Meanwhile, a noun closely follows the light verb by VOB as the direct object to supplement the semantic content of the verb. If the above structures are matched in the dependency trees, we would accurately extract CLVC phrases as a relation. To eliminate incoherent or uninformative extractions, we identify the combination of the verb and its direct object as a relationship expression, except for the preposition.

\begin{figure}
	\centerline{\includegraphics[width=0.85\textwidth]{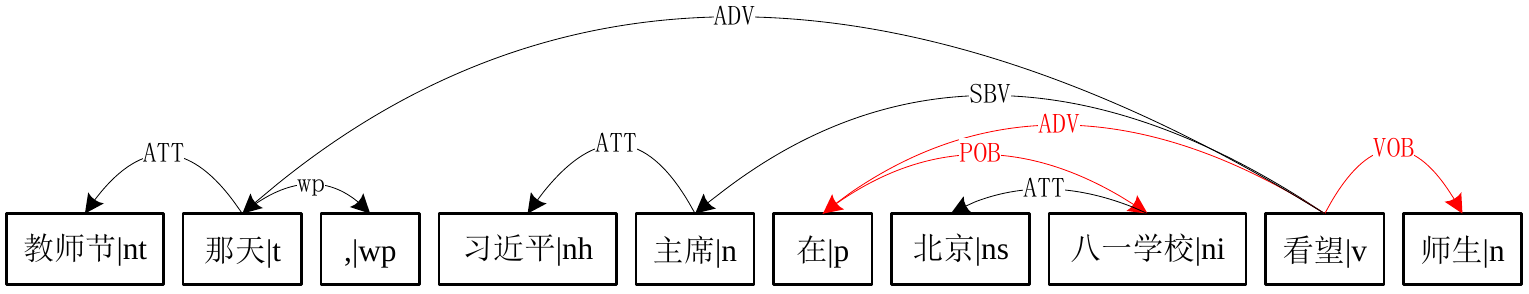}}
	\caption{The results of segments, POS-tagging and syntactic parsing for example sentence using to illustrate the Extend CLVC.  Note that we mark the key construction in red.}
	\label{fig:three1}
\end{figure}

Furthermore, a transitive notional verb can be used independently as a predicate. It also may be used in conjunction with a noun (direct object) and have a prepositional phrase as an adverbial modifier to modify the verb. For instance, ``\begin{CJK*}{UTF8}{gbsn}教师节那天\end{CJK*}(On Teachers' day), \begin{CJK*}{UTF8}{gbsn}习近平\end{CJK*}(Xi Jinping) \begin{CJK*}{UTF8}{gbsn}主席\end{CJK*}(president) \begin{CJK*}{UTF8}{gbsn}在\end{CJK*}(at) \begin{CJK*}{UTF8}{gbsn}北京八一学校\end{CJK*}(Beijing Bayi School) \begin{CJK*}{UTF8}{gbsn}看望\end{CJK*}(visited) \begin{CJK*}{UTF8}{gbsn}师生\end{CJK*}(teachers and students).", whose syntactic parsing is shown in Figure~\ref{fig:three1}. The sentence parsing satisfies the above description. Based on this analysis we can get the triple (\begin{CJK*}{UTF8}{gbsn}习近平\end{CJK*}(Xi Jinping), \begin{CJK*}{UTF8}{gbsn}看望\end{CJK*}(visited) \begin{CJK*}{UTF8}{gbsn}师生\end{CJK*}(teachers and students), \begin{CJK*}{UTF8}{gbsn}北京八一学校\end{CJK*}(Beijing Bayi School)). Although the notional verb is semantically independent, only the combination of predicates and nouns can express full textual semantics. Thus, we define this phenomenon as the {\bfseries Extended CLVC}.

ZORE \cite{Qiu2014} used a statistical metric to determine whether this kind of structure is a valid LVC to exclude the extend LVC. It identifies the condition as a four tuple (\begin{CJK*}{UTF8}{gbsn}习近平\end{CJK*}(Xi Jinping), \begin{CJK*}{UTF8}{gbsn}北京八一学校\end{CJK*}(Beijing Bayi School), \begin{CJK*}{UTF8}{gbsn}看望\end{CJK*}(visited), \begin{CJK*}{UTF8}{gbsn}师生\end{CJK*}(teachers and students)). By comparison, in this article, Three tuples are deliberately extracted by combining verbs and their direct objects as relational expressions.

\begin{definition}Intransitive Verb (IV) Phenomenon.\end{definition}
IV means that intransitive verb must be linked to it's patient by preposition, and the preposition may be on the left or right side of the verb.

According to whether the verbs can be directly with the object, verbs are divided into two types: transitive verb and intransitive verb. The object cannot directly follow the intransitive verb. In English, a preposition has to be added between an intransitive verb and object, such as ``born in", ``work at" and ``apologize to".

\begin{figure}[H]
	\centering
	\subfigure[]{
		\label{fig:three:a} 
		\includegraphics[width=0.7\textwidth]{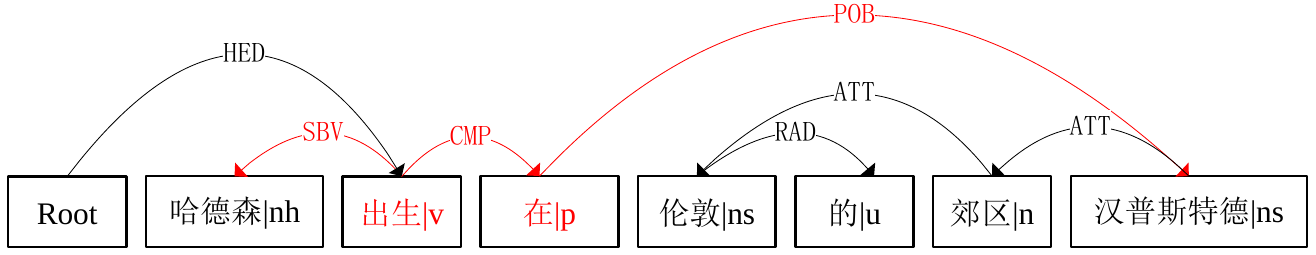}}
	\hspace{1in} \subfigure[]{
		\label{fig:three:b} 
		\includegraphics[width=0.7\textwidth]{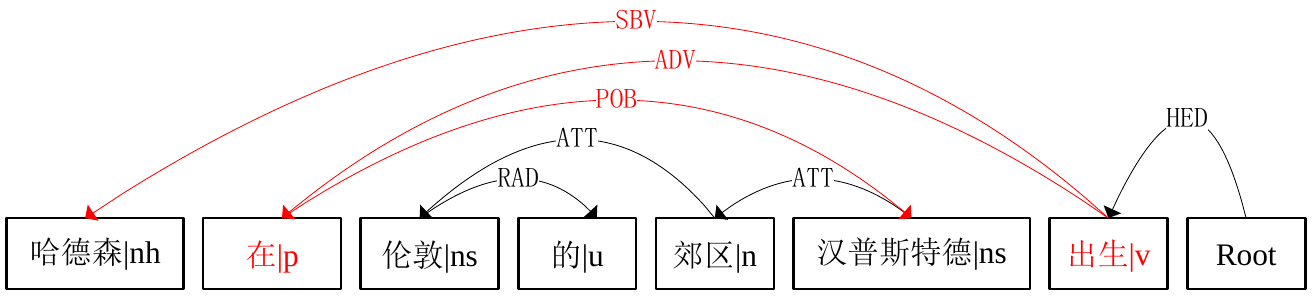}}
	\caption{The both graphs show the dependency trees of two kinds of Chinese translation. Note that we mark the key construction in red.}
	\label{fig:three} 
\end{figure}

REVERB handles the problem by syntactic constraints. The relationship phrase must be a contiguous sequence of words in the sentence, begin with a verb, end with a preposition. The sentence ``Hudson born in Hampstead, which is a suburb of London." can result in the relation phrase ``born in" which is an IV phenomenon. We can translate this sentence into Chinese by two ways: \begin{CJK*}{UTF8}{gbsn}哈德森\end{CJK*}(Hudson) \begin{CJK*}{UTF8}{gbsn}出生\end{CJK*}(born) \begin{CJK*}{UTF8}{gbsn}在\end{CJK*}(in) \begin{CJK*}{UTF8}{gbsn}伦敦\end{CJK*}(London) \begin{CJK*}{UTF8}{gbsn}的\end{CJK*}(de, an auxiliary word) \begin{CJK*}{UTF8}{gbsn}郊区\end{CJK*}(suburb) \begin{CJK*}{UTF8}{gbsn}汉普斯特德\end{CJK*}(Hampstead) or \begin{CJK*}{UTF8}{gbsn}哈德森\end{CJK*}(Hudson) \begin{CJK*}{UTF8}{gbsn}在\end{CJK*}(in) \begin{CJK*}{UTF8}{gbsn}伦敦\end{CJK*}(London) \begin{CJK*}{UTF8}{gbsn}的\end{CJK*}(de) \begin{CJK*}{UTF8}{gbsn}郊区\end{CJK*}(suburb) \begin{CJK*}{UTF8}{gbsn}汉普斯特德\end{CJK*}(Hampstead) \begin{CJK*}{UTF8}{gbsn}出生\end{CJK*}(born). The two kinds of Chinese translations are dependency-parsed and shown in Figure~\ref{fig:three}. Obviously, the preposition \begin{CJK*}{UTF8}{gbsn}在\end{CJK*}(in) can be on the left or right of the verb \begin{CJK*}{UTF8}{gbsn}出生\end{CJK*}(born) in Chinese.

Consistent with the principle of dealing with CLVC, dependency parsing benefits to exploring the IV, without worrying about the impact of the preposition's position. When the preposition is located at the left of the verb, the preposition depends on the verb by \emph{ADV}, and there is an entity as an object depending on preposition by \emph{POB}. Otherwise, by lying on the right of the verb, the preposition depends on the verb by \emph{CMP}, and the entity depends on the preposition by \emph{POB}. If the above structures are matched in the dependency trees, we would accurately extract IV phrases as a relation.

IV is treated as a CLVC in some studies e.g. \cite{Qiu2014}. We think such taxonomy is barely appropriate, as they own different linguistic sense. Although IV is similar to CLVC in that both need prepositions to construct sentences and can be treated similarly during actually extraction processing, intransitive verbs that have explicit semantics are almost distinguished from the light verbs.

\vspace{2ex}

We randomly collect 500 sentences from web news text, in which we manually extract 118 correct relations. Statistics show that relation triples that agree with the unique phenomena account for over 60\% of the total. Among them, NMCC has the highest frequency, accounting for approximately 40.37\%. Second, CLVC holds 14.64\%, and IVC has 8.78\%. These unique Chinese linguistic phenomena are universal. There will be an unfortunate effect on the final results if we ignore any of them.


\bibliographystyle{splncs04}
\bibliography{DSNF}
%
%
%
%

\end{document}